# Generating New Beliefs From Old*

**Fahiem Bacchus**
Computer Science Dept.
University of Waterloo
Waterloo, Ontario
Canada, N2L 3G1
fbacchus@logos.waterloo.edu

**Adam J. Grove**
NEC Research Institute
4 Independence Way
Princeton, NJ 08540
grove@research.nj.nec.com

**Joseph Y. Halpern**
IBM Almaden Research Center
650 Harry Road
San Jose, CA 95120–6099
halpern@almaden.ibm.com

**Daphne Koller**
Computer Science Division
University of California, Berkeley
Berkeley, CA 94720
daphne@cs.berkeley.edu

## Abstract

In previous work [BGHK92, BGHK93], we have studied the *random-worlds* approach—a particular (and quite powerful) method for generating degrees of belief (i.e., subjective probabilities) from a knowledge base consisting of objective (first-order, statistical, and default) information. But allowing a knowledge base to contain only objective information is sometimes limiting. We occasionally wish to include information about degrees of belief in the knowledge base as well, because there are contexts in which old beliefs represent important information that should influence new beliefs. In this paper, we describe three quite general techniques for extending a method that generates degrees of belief from objective information to one that can make use of degrees of belief as well. All of our techniques are based on well-known approaches, such as *cross-entropy*. We discuss general connections between the techniques and in particular show that, although conceptually and technically quite different, all of the techniques give the same answer when applied to the random-worlds method.

## 1 Introduction

When we examine the knowledge or information possessed by an agent, it is useful to distinguish between *subjective* and *objective* information. Objective information is information about the environment, whereas subjective information is information about the state of the agent's beliefs. For example, we might characterize the information of an agent travelling from San Francisco to New York as consisting of the objective information that the weather is warm in San Francisco, and the subjective information that the probability that the weather is warm in New York is 0.2. The important thing to notice here is that although we can in principle determine if the agent's objective information is correct (by examining what is actually the case in its environment), we cannot so easily say that its subjective beliefs are correct. The truth or falsity of these pieces of information is not determined by the state of the environment.

Although subjective information could take many different forms, we will concentrate here on *degrees of belief*. These are probabilities that are assigned to formulas expressing objective assertions. For example, the assertion "the weather is warm in New York" is an objective one: it is either true or false in the agent's environment. But when we assign a degree of belief to this assertion, as above, we obtain a subjective assertion: it becomes a statement about the state of the agent's beliefs. In the context of probability theory the distinction between subjective and objective can appear somewhat subtle, because some form of objective information (such as proportions or frequencies) obey the laws of probability, just as do degrees of belief. Yet the distinction can be a significant one if we want to use or interpret a probabilistic theory correctly. Carnap's work [Car50] is noteworthy for its careful distinction between, and study of, both statistical probabilities, which are objective, and degree of belief probabilities, which are subjective.

In order to understand this distinction, it is useful to provide a formal semantics for degrees of belief that captures the difference between them and objective information. As demonstrated by Halpern [Hal90], a natural, and very general, way to give a semantics to degrees of belief is by defining a probability distribution over a set of possible worlds.[1] The degree of belief in a formula $\varphi$ is then the probability of the set of worlds where $\varphi$ is true. In this framework we can characterize objective information as consisting of assertions (expressed as formulas) that can be assigned a truth value by a *single* world. For example, in any given world Tweety the bird does or does not fly. Hence, the formula *Fly(Tweety)* is objective. Statistical assertions such as $||Fly(x)|Bird(x)||_x \approx 0.8$, read "approximately 80% of birds fly", are also objective. On the other hand, $\Pr(Fly(Tweety)) = 0.8$, expressing the assertion that the agent's degree of belief in Tweety flying is 0.8, is not objective, as its truth is determined by whether or not the probability of the *set* of worlds where Tweety flies is 0.8.

---

*This research has been supported in part by the Canadian Government through their NSERC and IRIS programs, by the Air Force Office of Scientific Research (AFSC) under Contract F49620-91-C-0080, and by a University of California President's Postdoctoral Fellowship. The United States Government is authorized to reproduce and distribute reprints for governmental purposes.

[1]Conceptually, this notion of world is just as in classical "possible-worlds semantics": a complete picture or description of the way the world might be. Formally, we take a world to be an interpretation (model) for first-order logic.

Although we cannot easily characterize an agent's degrees of beliefs as being correct or incorrect, it is nevertheless clear that these beliefs should have some relation to objective reality. One way of guaranteeing this is to actually generate them from the objective information available to the agent. Several ways of doing this have been considered in the literature; for example, [BGHK92, PV92] each discuss several possibilities. The approaches in [BGHK92] are based in a very natural way on the semantics described above. Assume we have a (prior) probability distribution over some set of worlds. We can then generate degrees of belief from an objective knowledge base KB by using standard Bayesian conditioning; to the formula $\varphi$ we assign as its degree of belief the conditional probability of $\varphi$ given KB. In [BGHK92] we considered three particular choices for a prior, and investigated the properties of the resulting inductive inference systems. In [BGHK93] we concentrated on the simplest of these methods—the *random-worlds* method—whose choice of prior is essentially the uniform prior over the set of possible worlds.

More precisely, suppose we restrict our attention to worlds (i.e., interpretations of an appropriate vocabulary for first-order logic) with the domain $\{1, \ldots, N\}$. Assuming we have a finite vocabulary, there will be only finitely many such worlds. Random worlds takes as the set of worlds all of these worlds, and uses perhaps the simplest probability distribution over them—the uniform distribution—thus assuming that each of the worlds is equally likely. This gives a prior distribution on the set of possible worlds. We can now induce a degree of belief in $\varphi$ given KB by using the conditional probability of $\varphi$ given KB with respect to this uniform distribution. It is easy to see that the degree of belief in $\varphi$ given KB is then simply the fraction of possible worlds satisfying KB that also satisfy $\varphi$. In general, however, we do not know the domain size $N$; we know only that it is typically large. We can therefore approximate the degree of belief for the true but unknown $N$ by computing the limiting value of this degree of belief as $N$ grows large. This limiting value (if it exists, which it may not) is denoted $\Pr_\infty^w(\varphi|KB)$, and it is what the random-world method takes to be the degree of belief in $\varphi$ given KB. In [BGHK93], we showed that this method possesses a number of attractive properties, such as a preference for more specific information and the ability to ignore irrelevant information.

The random-worlds method can generate degrees of belief from rich knowledge bases that may contain first-order, statistical, and default information. However, as with any conditioning process, is limited to dealing with *objective* information. When we add subjective formulas to KB, we can no longer simply condition on KB: the conditioning process eliminates those worlds inconsistent with our information, while the truth of a subjective formula cannot be determined by a single world.[2] Hence, we would like to extend the random-worlds method so as to enable it to deal with both objective and subjective information.

Why do we want to take into account subjective beliefs? There are a number of situations where this seems to make sense. For example, suppose a birdwatcher is interested in a domain of birds, and has an objective knowledge base $KB_{bird}$ consisting of the statistical information

$$||Cardinal(x)|\neg Red(x)||_x \approx 0.1 \ \wedge$$
$$||Cardinal(x)|Red(x)||_x \approx 0.7.$$

Now the birdwatcher catches a glimpse of a bird $b$ flying by that seems to be red. The birdwatcher is trying to decide if $b$ is a cardinal. By the results of [BGHK93], if the birdwatcher assumes that the bird is not red, random-worlds gives $\Pr_\infty^w(Cardinal(b)|KB_{bird} \wedge \neg Red(b)) = 0.1$. On the other hand, if she assumes that the bird is red, we get $\Pr_\infty^w(Cardinal(b)|KB_{bird} \wedge Red(b)) = 0.7$. But it does not seem appropriate for her to do either; rather we would like to be able to generate a degree of belief in $Cardinal(b)$ that takes into account the birdwatcher's degree of belief in $Red(b)$. For example, if this degree of belief is 0.8, then we would like to use a knowledge base such as $KB_{bird} \wedge Pr(Red(b)) = 0.8$. It seems reasonable to expect that the resulting degree of belief in $Cardinal(b)$ would then be somewhere between the two extremes of 0.7 and 0.1.

As another example, suppose we have reason to believe that two sensors are independent. For simplicity, suppose the sensors measure temperature, and report it to be either high, $h$, or low, $l$. We can imagine three unary predicates: $S1(x)$, indicating that sensor 1 reports the value $x$; $S2(x)$, a similar predicate for sensor 2; and $Actual(x)$, indicating that the actual temperature is $x$. That the sensors are independent (given the actual value) can be represented by the conjunction over all choices for $x$, $x'$, and $x''$ in $\{l, h\}$ of:

$$\Pr(S1(x') \wedge S2(x'')|Actual(x))$$
$$= \Pr(S1(x')|Actual(x)) \times \Pr(S2(x'')|Actual(x)).$$

It could be that we have determined that the sensors are independent through the observation of a number of test readings. Such empirical evidence could be summarized by a statistical assertion and thus added to our knowledge base without requiring a degree of belief statement like the above. However, this is not the normal situation. Rather, we are more likely to have based our belief in independence on other information, such as our beliefs about causality. For example, the sensors may have been built by different manufacturers. In this case, it seems most reasonable to represent this kind of information using an assertion about degrees of belief.

How, then, can we incorporate information about degrees of belief into the random-worlds framework? More generally, given any *inference process*[3] —i.e., a method for generating degrees of belief from objective information—we would

---

[2]In the context of random worlds (and in other cases where the degrees of belief are determined using a prior on the set of

worlds), this problem can be viewed as an instance of the general problem of conditioning a distribution on uncertain evidence.

[3]The term "inference process" is taken from Paris and Vencovska [PV89]. Our framework is slightly different from theirs, but we think this usage of the term is consistent with their intent.

like to extend it so that it can also deal with subjective information. This is an issue that has received some attention recently [PV92, Jae94b, Jae94a]. We discuss three techniques here, and consider their application in the specific context of random worlds. As we shall see, all of our techniques are very closely based on well-known ideas in the literature. Two make use of *cross-entropy*, while the third is a generalization of a method considered by Paris and Vencovska [PV92]. They are conceptually and formally distinct, yet there are some interesting connections between them. In particular, in the context of random-worlds they generally yield the same answers (where the comparison makes sense; the various methods have different ranges of applicability). Many of the results we discuss are, in general terms if not in specific details, already known. Nevertheless, their combination is quite interesting.

We now describe the three methods in a little more detail. The first method we examine is perhaps the simplest to explain. We consider it first in the context of random worlds. Fix $N$. Random worlds considers all of the worlds that have domain $\{1, \ldots, N\}$, and assumes they are equally likely, which seems reasonable in the absence of information to the contrary. But now suppose that we have a degree of belief such as $\Pr(Red(b)) = 0.8$. In this case it is no longer reasonable to assume that all worlds are equally likely; our knowledge base tells us that the worlds where $b$ is red are more likely than the worlds where $b$ is not red. Nevertheless, there is a straightforward way of incorporating this information. Rather than taking all worlds to be equally likely, we divide the worlds into two sets: those which satisfy $Red(b)$ and those which satisfy $\neg Red(b)$. Our beliefs require that the first set have probability 0.8 and the second probability 0.2. But otherwise we can make the worlds within each set equally likely. This is consistent with the random worlds approach of making all worlds equally likely. Intuitively, we are considering the probability distribution on the worlds that is as close as possible to our original uniform distribution subject to the constraint that the set of worlds where $Red(b)$ holds should have probability 0.8.

What do we do if we have an inference process other than random worlds? As long as it also proceeds by generating a prior on a set of possible worlds and then conditioning, we can deal with at least this example. We simply use the prior generated by the method to assign relative weights to the worlds in the sets determined by $Red(b)$ and $\neg Red(b)$, and then scale these weights within each set so that the sets are assigned probability 0.8 and 0.2 respectively. (Readers familiar with Jeffrey's rule [Jef92] will realize that this is essentially an application of that rule.) Again, intuitively, we are considering the distribution closest to the original prior that gives the set of worlds satisfying $Red(b)$ probability 0.8.

Unfortunately, the knowledge base is rarely this simple. Our degrees of belief often place complex constraints on the probability distribution over possible worlds. Nevertheless, we would like to maintain the intuition that we are considering the distribution "closest" to the original prior that satisfies the constraints imposed by the KB. But how do we determine the "closest" distribution? One way is by using *cross-entropy* [KL51]. Given two probability distributions $\mu$ and $\mu'$, the cross-entropy of $\mu'$ relative to $\mu$, denoted $C(\mu', \mu)$, is a measure of how "far" $\mu'$ is from $\mu$ [SJ80, Sho86]. Given an inference method that generates a prior and a set of constraints determined by the KB, we can then find the distribution on worlds satisfying the constraints that minimizes cross-entropy relative to the prior, and then use this new distribution to compute degrees of belief. We call this method CEW (for *cross-entropy on worlds*).

The next method we consider also uses cross-entropy, but in a completely different way. Suppose we have the (objective) knowledge base $\mathrm{KB}_{bird}$ given above, and a separate "belief base" $\mathrm{BB}_{bird} = (\Pr(Red(b)) = 0.8)$. As we suggested, if the birdwatcher were sure that $b$ was red, random worlds would give a degree of belief of 0.7 in $Cardinal(b)$; similarly, if she were sure that $b$ was not red, random worlds would give 0.1. Given that her degree of belief in $Red(b)$ is 0.8, it seems reasonable to assign a degree of belief of $0.8 \times 0.7 + 0.2 \times 0.1$ to $Cardinal(b)$. In fact, if we consider any inference process $I$ (not necessarily one that generates a prior probability on possible worlds), it seems reasonable to define

$$I(Cardinal(b)|\mathrm{KB}_{bird} \wedge \mathrm{BB}_{bird})$$
$$= 0.8 \times I(Cardinal(b)|\mathrm{KB}_{bird} \wedge Red(b))$$
$$+ 0.2 \times I(Cardinal(b)|\mathrm{KB}_{bird} \wedge \neg Red(b)).$$

More generally, we might hope that given an inference process $I$ and a knowledge base of the form $\mathrm{KB} \wedge \mathrm{BB}$, we can generate from it a collection of objective knowledge bases $\mathrm{KB}_1, \ldots, \mathrm{KB}_m$ such that $I(\varphi|\mathrm{KB} \wedge \mathrm{BB})$ is a weighted average of $I(\varphi|\mathrm{KB}_1), \ldots, I(\varphi|\mathrm{KB}_m)$, as in the example. In general, however, achieving this in a reasonable fashion is not so easy. Consider the belief base $\mathrm{BB}'_{bird} = (\Pr(Red(b)) = 0.8) \wedge (\Pr(Small(b)) = 0.6)$. In this case, we would like to define $I(Cardinal(b)|\mathrm{KB}_{bird} \wedge \mathrm{BB}'_{bird})$ using a weighted average of $I(Cardinal(b)|\mathrm{KB}_{bird} \wedge Red(b) \wedge Small(b))$, $I(Cardinal(b)|\mathrm{KB}_{bird} \wedge Red(b) \wedge \neg Small(b))$, etc. As in the simple example, it seems reasonable to take the weight of the term $I(Cardinal(b)|\mathrm{KB}_{bird} \wedge Red(b) \wedge Small(b))$ to be the degree of belief in $Red(b) \wedge Small(b)$. Unfortunately, while $\mathrm{BB}'_{bird}$ tells us the degree of belief in $Red(b)$ and $Small(b)$ separately, it does not give us a degree of belief for their conjunction. A superficially plausible heuristic would be to assume that $Red(b)$ and $Small(b)$ are independent, and thus assign degree of belief $0.8 \times 0.6$ to their conjunction. While this seems reasonable in this case, at other times it is completely inappropriate. For example, if our knowledge base asserts that all small things are red, then $Red(b)$ and $Small(b)$ cannot be independent, and we should clearly take the degree of belief in $Red(b) \wedge Small(b)$ to be the same as the degree of belief in $Small(b)$, namely, 0.6. In general, our new degree of belief for the formula $Red(b) \wedge Small(b)$ may depend not only on the new degrees of belief for the two conjuncts, but also on our old degree of belief $I(Red(b) \wedge Small(b)|\mathrm{KB}_{bird})$. One reasonable approach to computing these degrees of belief is to make the *smallest* change possible to achieve consistency with the belief base. Here, as before, *cross-entropy* is a useful tool. Indeed, as we shall show, there is a way of applying cross-entropy in this context to give us a general approach. We call this method CEF, for *cross-entropy on formulas*. Although both CEW and CEF use cross-entropy, they use it in conceptually different ways. As the names suggest, CEW uses cross-entropy to compare two probability distributions over possible worlds, while CEF uses it to compare two probability distributions over formulas. On the other hand, any probability distribution on worlds generates a probability distribution on formulas in the obvious way (the probability of a formula is the probability of the set of worlds where it is true), and so we can use a well-known property of the cross-entropy function to observe that the two approaches are in fact equivalent when they can both be applied.

It is worth noting that the two approaches are actually incomparable in their scope of application. Because CEF is not restricted to inference processes that generate a prior probability on a set of possible worlds, it can be applied to more inference processes than CEW. On the other hand, CEW is applicable to arbitrary KB's while, as we shall see, for CEF to apply we need to make more restrictions on the form of the KB.

In this paper, we focus on two instantiations of CEF. The first applies it to the random-worlds method. The second applies it to a variant of the maximum-entropy approach used by Paris and Vencovska [PV89] (and similar in spirit to the method used by Jaeger [Jae94b]), which we henceforth call the *ME (inference) process*. Using results of [GHK92, PV89], we prove that these two instantiations are equivalent.

The third method we consider also applies only to certain types of inference processes. In particular, it takes as its basic intuition that all degrees of belief must ultimately be the result of some statistical process. Hence, it requires an inference process that can generate degrees of belief from statistics, like random-worlds. Suppose we have the belief $Pr(Red(b)) = 0.8$. If we view this belief as having arisen from some statistical sampling process, then we can regard it as an abbreviation for statistical information about the class of individuals who are "just like $b$". For example, say that we get only a quick glance at $b$, so we are not certain it is red. The above assertion could be construed as being an abbreviated way of saying that 80% of the objects that give a similar sense perception are red. To capture this idea formally we can view $b$ as a member of a small set of (possibly fictional) individuals $S$ that are "just like $b$" to the best of our knowledge, and assume that our degrees of belief about $b$ actually represents the statistical information about $S$: $||Red(x)|S(x)||_x \approx 0.8$. Once all degree of belief assertions have been converted into statistical assertions, we can then apply any method for inferring degrees of belief from statistical knowledge bases. We call this the RS method (for *representative set*). The general intuition for this method goes back to statistical mechanics [Lan80]. It was also defined (independently it seems) by Paris and Vencovska [PV92]; we follow their presentation here.

Paris and Vencovska showed that the RS method and the CEF method agree when applied to their version of the ME process. Using results of [GHK92, PV89], we can show that the methods also agree when applied to our version of the ME process and when applied to random worlds. Putting the results together, we can show that all these methods—CEW, CEF, and RS—agree when applied to random worlds and, in fact, CEW and CEF agree in general. In addition, the resulting extension of random worlds agrees with the approach obtained when we apply CEF and RS to the ME process.

The rest of this paper is organized as follows. In the next section we review the formal model of [Hal90] for degrees of belief and statistical information, and some material from [BGHK93] regarding the random-worlds method. We give the formal definitions of the three methods we consider in Section 3, and discuss their equivalence. In passing, we also discuss the connection to Jeffrey's rule, which is another very well known method of updating by uncertain information. We conclude in Section 4 with some discussion of computational issues and possible generalizations of these approaches.

## 2   Technical preliminaries

### 2.1   A first-order logic of probability

In [Hal90], a logic is presented that allows us to represent and reason with both statistical information and degrees of belief. We briefly review the relevant material here. We start with a standard first-order language over a finite vocabulary $\Phi$, and augment it with *proportion expressions* and *belief expressions*. A *basic proportion expression* has the form $||\psi(x)|\theta(x)||_x$ and denotes the proportion of domain elements satisfying $\psi$ from among those elements satisfying $\theta$. (We take $||\psi(x)||_x$ to be an abbreviation for $||\psi(x)|true(x)||_x$.) On the other hand, a *basic belief expression* has the form $Pr(\psi|\theta)$ and denotes the agent's degree of belief in $\psi$ given $\theta$. The set of proportion (resp. belief) expressions is formed by adding the rational numbers to the set of basic proportion (resp. belief) expressions and then closing off under addition and multiplication.

We compare two proportion expressions using the approximate connective $\preceq$ ("approximately less than or equal"); the result is a *proportion formula*. We use $\xi \approx \xi'$ as an abbreviation for $(\xi \preceq \xi') \wedge (\xi' \preceq \xi)$. Thus, for example, we can express the statement "90% of birds fly" using the proportion formula $||Fly(x)|Bird(x)||_x \approx 0.9$.[4] We compare two belief expressions using standard $\leq$; the result is a *basic belief formula*. For example, $Pr(Red(b)) \leq 0.8$ is a basic belief formula. (Of course, $Pr(Red(b)) = 0.8$ can be expressed as the obvious conjunction.) In the full language $\mathcal{L}$ we allow arbitrary first-order quantification and nesting of belief and proportion formulas. For example, complex formulas like $Pr(\forall x(||Knows(x, y)||_y \preceq 0.3)) \leq 0.5$ are in $\mathcal{L}$.

---

[4]We remark that in [Hal90] there was no use of approximate equality ($\approx$). We use it here since, as argued in [BGHK93], its use is crucial in our intended applications. On the other hand, in [BGHK93], we used a whole family of approximate equality functions of the form $\approx_i$, $i = 1, 2, 3, \ldots$. To simplify the presentation, we use only one here.

We will also be interested in various sublanguages of $\mathcal{L}$. A formula in which the "Pr" operator does not appear is an *objective formula*. Such formulas are assigned truth values by single worlds. The sublanguage restricted to objective formulas is denoted by $\mathcal{L}^{obj}$. The standard random-worlds method is restricted to knowledge bases expressed in $\mathcal{L}^{obj}$. The set of belief formulas, $\mathcal{L}^{bel}$, is formed by starting with basic belief formulas and closing off under conjunction, negation, and first-order quantification. In contrast to objective formulas, the truth value of a belief formula is completely independent of the world where it is evaluated. A *flat formula* is a Boolean combination of belief formulas, such that in each belief expression $\Pr(\varphi)$, the formula $\varphi$ is a closed (i.e., containing no free variables) objective formula. (Hence we have no nesting of "Pr" in flat formulas nor any "quantifying in".) Let $\mathcal{L}^{flat}$ be the language consisting of the flat formulas.

To give semantics to both proportion formulas and belief formulas, we use a special case of what were called in [Hal90] *type-3 structures*. In particular, we consider type-3 structures of the form $(\mathcal{W}_N, \mu)$, where $\mathcal{W}_N$ consists of all worlds (first-order models) with domain $\{1, \ldots, N\}$ over the vocabulary $\Phi$, and $\mu$ is a probability distribution over $\mathcal{W}_N$.[5] Given a structure and a world in that structure, we evaluate a proportion expression $||\psi(x)||\theta(x)||_x$ as the fraction of domain elements satisfying $\psi(x)$ among those satisfying $\theta(x)$. We evaluate a belief formula using our probability distribution over the set of possible worlds. More precisely, given a structure $M = (\mathcal{W}_N, \mu)$, a world $w \in \mathcal{W}_N$, a tolerance $\tau \in (0, 1]$ (used to interpret $\approx$ and $\preceq$), and a valuation $V$ (used to interpret the free variables), we associate with each formula a truth value and with each belief expression or proportion expression $\zeta$ a number $[\zeta]_{M,w,V,\tau}$. We give a few representative clauses here:

- If $\zeta$ is the proportion expression $||\varphi(x)|\psi(x)||_x$, then $[\zeta]_{M,w,V,\tau}$ is the number of domain elements in $w$ satisfying $\varphi \wedge \psi$ divided by the number satisfying $\psi$. (Note that these numbers may depend on $w$.) We take this fraction to be 1 if no domain elements satisfies $\psi$.

- If $\zeta$ is the belief expression $\Pr(\varphi|\psi)$, then

$$[\zeta]_{M,w,V,\tau} = \frac{\mu\{w' : (M, w', V, \tau) \models \varphi \wedge \psi\}}{\mu\{w' : (M, w', V, \tau) \models \psi\}}.$$

  Again, we take this to be 1 if the denominator is 0.

- If $\zeta$ and $\zeta'$ are two proportion expressions, then $(M, w, \tau, V) \models \zeta \preceq \zeta'$ iff

$$[\zeta]_{M,w,\tau,V} \leq [\zeta']_{M,w,\tau,V} + \tau.$$

  That is, approximate less than or equal allows a tolerance of $\tau$.

Notice that if $\zeta$ is a belief expression, then its value is independent of the world $w$. Moreover, if it is closed then its value is independent of the valuation $V$. Thus, we can write $[\zeta]_{M,\tau}$ in this case. Similarly, if $\varphi \in \mathcal{L}^{bel}$ is a closed

---

[5] In general, type-3 structures additionally allow for a distribution over the domain (in this case, $\{1, \ldots, N\}$). Here, we always use the uniform distribution over the domain.

---

belief formula, its truth depends only on $M$ and $\tau$, so we can write $(M, \tau) \models \varphi$ in this case.

## 2.2 The random-worlds method

Given these semantics, the random-worlds method is now easy to describe. Suppose we have a KB of objective formulas, and we want to assign a degree of belief to a formula $\varphi$. Let $\mu_N^u$ be the uniform distribution over $\mathcal{W}_N$, and let $M_N^u = (\mathcal{W}_N, \mu_N^u)$. Let $\Pr_N^{\tau,rw}(\varphi|KB) = [\Pr(\varphi|KB)]_{M_N^u,\tau}$. Typically, we know only that $N$ is large and that $\tau$ is small. Hence, we approximate the value for the true $N$ and $\tau$ by defining

$$\Pr_\infty^{rw}(\varphi|KB) = \lim_{\tau \to 0} \lim_{N \to \infty} \Pr_N^{\tau,rw}(\varphi|KB),$$

assuming the limit exists. $\Pr_\infty^{rw}(\varphi|KB)$ is the degree of belief in $\varphi$ given KB according to the random-worlds method.

## 2.3 Maximum entropy and cross-entropy

The *entropy* of a probability distribution $\mu$ over a finite space $\Omega$ is $-\sum_{\omega \in \Omega} \mu(\omega) \ln(\mu(\omega))$. It has been argued [Jay78] that entropy measures the amount of "information" in a probability distribution, in the sense of information theory. The uniform distribution has the maximum possible entropy. In general, given some constraints on the probability distributions, the distribution with maximum entropy that satisfies the constraints can be viewed as the one that incorporates the least additional information above and beyond the constraints.

The related cross-entropy function measures the additional information gained by moving from one distribution $\mu$ to another distribution $\mu'$:

$$C(\mu', \mu) = \sum_{\omega \in \Omega} \mu'(\omega) \ln \frac{\mu'(\omega)}{\mu(\omega)}.$$

Various arguments have been presented showing that cross-entropy measures how close one probability distribution is to another [SJ80, Sho86]. Thus, given a prior distribution $\mu$ and a set $S$ of additional constraints, we are typically interested in the unique distribution $\mu'$ that satisfies $S$ and minimizes $C(\mu', \mu)$. It is well-known that a sufficient condition for such a unique distribution to exist is that the set of distributions satisfying $S$ form a convex set, and that there be at least one distribution $\mu''$ satisfying $S$ such that $C(\mu'', \mu)$ is finite. These conditions often hold in practice.

## 3 The three methods

### 3.1 CEW

As we mentioned in the introduction, our first method, CEW, assumes as input an inference process $I$ that proceeds by generating a prior $\mu_I$ on a set of possible worlds $\mathcal{W}_I$ and then conditioning on the objective information. Given such an inference process $I$, a knowledge base KB (that can contain subjective information) and an objective formula $\varphi$, we wish to compute $CEW(I)(\varphi|KB)$, where $CEW(I)$ is a new degree of belief generator that can handle knowledge bases that can include subjective information.

We say that an inference process $I$ is *world-based* if there is some structure $M_I = (\mathcal{W}_I, \mu_I)$ and a tolerance $\tau$ such that $I(\varphi|\text{KB}) = [\Pr(\varphi|\text{KB})]_{M_I, \tau}$. Notice that $\Pr_N^{\tau, \text{rw}}$ is world-based for each $N$ (where the structure corresponding to $\Pr_N^{\tau, \text{rw}}$ is $M_N^u$). $\Pr_\infty^{\text{rw}}$, on the other hand, is not world-based; we return to this point shortly.

Given a world-based inference process $I$, we define $CEW(I)$ as follows: Given a knowledge base KB which can be an arbitrary formula in the full language $\mathcal{L}$, let $\mu_I^{\text{KB}}$ be the probability distribution on $\mathcal{W}_I$ such that $C(\mu_I^{\text{KB}}, \mu_I)$ is minimized (if a unique such distribution exists) among all distributions $\mu'$ such that $(\mathcal{W}_I, \mu', \tau) \models \Pr(\text{KB}) = 1$. Intuitively, $\mu_I^{\text{KB}}$ is the probability distribution closest to the prior $\mu_I$ that gives KB probability 1. Let $M_I^{\text{KB}} = (\mathcal{W}_I, \mu_I^{\text{KB}})$. We can then define $CEW(I)(\varphi|\text{KB}) = [\Pr(\varphi)]_{M_I^{\text{KB}}, \tau}$.

The first thing to observe is that if KB is objective, then standard properties of cross-entropy can be used to show that $\mu_I^{\text{KB}}$ is the conditional distribution $\mu_I(\cdot|\text{KB})$. We thus immediately get:

**Proposition 3.1:** *If* KB *is objective, then* $CEW(I)(\varphi|\text{KB}) = I(\varphi|\text{KB})$. *Thus,* $CEW(I)$ *is a true extension of* $I$.

Another important property of CEW follows from the well-known fact that cross-entropy generalizes *Jeffrey's rule* [Jef92]. Standard probability theory tells us that if we start with a probability function $\mu$ and observe that event $E$ holds, we should update to the conditional probability function $\mu(\cdot|E)$. Jeffrey's rule is meant to deal with the possibility that rather than getting certain information, we only get partial information, such as that $E$ holds with probability $\alpha$. Jeffrey's rule suggests that in this case, we should update to the probability function $\mu'$ such that

$$\mu'(A) = \alpha\mu(A|E) + (1 - \alpha)\mu(A|\overline{E}),$$

where $\overline{E}$ denotes the complement of $E$. This rule uniformly rescales the probabilities within $E$ and (separately) those within $\overline{E}$ so as to satisfy the constraint $\Pr(E) = \alpha$. Clearly, if $\alpha = 1$, then $\mu'$ is just the conditional probability $\mu(\cdot|E)$.

This rule can be generalized in a straightforward fashion. If we are given a family of mutually exclusive and exhaustive events $E_1, \ldots, E_k$ with desired new probabilities $\alpha_1, \ldots, \alpha_k$ (necessarily $\sum_i \alpha_i = 1$), then we can define:

$$\mu'(A) = \alpha_1\mu(A|E_1) + \cdots + \alpha_K\mu(A|E_k).$$

Suppose our knowledge base has the form $(\Pr(\varphi_1) = \alpha_1) \wedge \cdots \wedge (\Pr(\varphi_k) = \alpha_k)$, where the $\varphi_i$'s are mutually exclusive and exhaustive objective formulas and $\alpha_1 + \cdots + \alpha_k = 1$. The formulas $\varphi_1, \ldots, \varphi_k$ correspond to mutually exclusive and exhaustive events. Thus, Jeffrey's rule would suggest that to compute the degree of belief in $\varphi$ given this knowledge base, we should compute the degree of belief in $\varphi$ given each of the $\varphi_i$ separately, and then take the linear combination. Using the fact that cross-entropy generalizes Jeffrey's rule, it is immediate that CEW in fact does this.

**Proposition 3.2:** *Suppose that* $I$ *is a world-based inference process and that* $\text{KB}'$ *is of the form* KB $\wedge$ BB, *where* KB *is objective and* BB *has the form* $(\Pr(\varphi_1) = \alpha_1) \wedge \cdots \wedge (\Pr(\varphi_k) = \alpha_k)$, *where the* $\varphi_i$'s *are mutually exclusive and exhaustive objective formulas and* $\alpha_1 + \cdots + \alpha_k = 1$. *Then*

$$CEW(I)(\varphi|\text{KB}') = \sum_{i=1}^{k} \alpha_i I(\varphi|\text{KB} \wedge \varphi_i).$$

As we observed above, CEW as stated does not apply directly to the random-worlds method $\Pr_\infty^{\text{rw}}$, since it is not world-based. It is, however, the limit of world-based methods. (This is also true for the other methods considered in [BGHK92].) We can easily extend CEW so that it applies to limits of world-based methods by taking limits in the obvious way. In particular, we define

$$CEW(\Pr_\infty^{\text{rw}})(\varphi|\text{KB}) = \lim_{\tau \to 0} \lim_{N \to \infty} CEW(\Pr_N^{\tau, \text{rw}})(\varphi|\text{KB}),$$

provided the limit exists. For convenience, we abbreviate $CEW(\Pr_\infty^{\text{rw}})$ as $\Pr_\infty^{\text{CEW}}$.

It is interesting to note that the distribution defined by $CEW(\Pr_N^{\text{rw}})$ is the distribution of maximum entropy that satisfies the constraint $\Pr(\text{KB}) = 1$. This follows from the observation that the distribution that minimizes the cross-entropy from the uniform distribution among those distributions satisfying some constraints $S$, is exactly the distribution of maximum entropy satisfying $S$.[6] This maximum-entropy characterization demonstrates that $\Pr_\infty^{\text{CEW}}$ extends random worlds by making the probabilities of the possible worlds "as equal as possible" given the constraints.

### 3.2 CEF

Paris and Vencovska [PV89] consider inferences processes that are not world-based, so CEW cannot be applied to them. The method CEF we now define applies to arbitrary inference processes, but requires that the knowledge base be of a restricted form. For the remainder of this section, we assume that the knowledge base has the form KB $\wedge$ BB, where KB is an objective formula and BB (which we call the belief base) is in $\mathcal{L}^{fixt}$.

First, suppose for simplicity that BB is of the form $\Pr(\psi_1) = \beta_1 \wedge \cdots \wedge \Pr(\psi_k) = \beta_k$. If the $\psi_i$'s were mutually exclusive, then we could define $CEF(I)(\varphi|\text{BB})$ so that Proposition 3.2 held. But what if the $\psi_i$'s are not mutually exclusive?

Consider the $K = 2^k$ *atoms* over $\psi_1, \ldots, \psi_k$, i.e., those conjunctions of the form $\psi_1' \wedge \ldots \wedge \psi_k'$, where each $\psi_i'$ is either $\psi_i$ or $\neg\psi_i$. Atoms are always mutually exclusive and exhaustive; so, if we could find appropriate degrees of belief for these atoms, we could again define things so that Proposition 3.2 holds. A simple way of doing this would be to assume that, after conditioning, the assertions $\psi_i$ are independent. But, as we observed in the introduction, assuming independence is inappropriate in general.

---

[6] We remark that in [GHK92, PV89] a connection was established between random worlds and maximum entropy. Here maximum entropy is playing a different role. It is being used here to extend random worlds rather than to characterize properties of random worlds as in [GHK92, PV89].

Our solution is to first employ cross-entropy to find appropriate probabilities for these atoms. We proceed as follows. Suppose $I$ is an arbitrary inference process, $BB \in \mathcal{L}^{flat}$, and $\psi_1, \ldots, \psi_k$ are the formulas that appear in subexpressions of the form $\Pr(\psi)$ in BB. We form the $K = 2^k$ atoms generated by the $\psi_i$, denoting them by $A_1, \ldots, A_K$. Consider the probability $\mu$ defined on the space of atoms via $\mu(A_j) = I(A_j | KB)$.[7] There is an obvious way of defining whether the formula BB is satisfied by a probability distribution on the atoms $A_1, \ldots, A_k$ (we defer the formal details to the full paper), but in general BB will not be satisfied by the distribution $\mu$. For a simple example, if we take the inference procedure to be random worlds and consider the knowledge base $KB_{bird} \wedge (\Pr(Red(b)) = 0.8)$ from the introduction, it turns out that $\Pr_\infty^{rw}(Red(b) | KB_{bird})$ is around 0.57. Clearly, the distribution $\mu$ such that $\mu(Red(b))$ is around 0.57 does not satisfy the constraint $\Pr(Red(b)) = 0.8$. Let $\mu'$ be the probability distribution over the atoms that minimizes cross-entropy relative to $\mu$ among those that satisfy BB, provided there is a unique such distribution. We then define

$$CEF(I)(\varphi | KB \wedge BB) =$$
$$\mu'(A_1) I(\varphi | KB \wedge A_1) + \cdots + \mu'(A_K) I(\varphi | KB \wedge A_K).$$

It is immediate from the definition that $CEF(I)$ extends $I$. Formally, we have

**Proposition 3.3:** If $KB, \varphi \in \mathcal{L}^{obj}$, then $CEF(I)(\varphi | KB) = I(\varphi | KB)$.

Both CEW and CEF use cross-entropy. However, the two applications are quite different. In the case of CEW, we apply cross-entropy with respect to probability distributions over possible worlds, whereas with CEF, we apply it to probability distributions over formulas. Nevertheless, as we mentioned in the introduction, there is a tight connection between the approaches, since any probability distribution over worlds defines a probability distribution over formulas. In fact the following equivalence can be proved, using simple properties of the cross-entropy function.

**Theorem 3.4:** *Suppose $I$ is a world-based inference process,* $KB, \varphi \in \mathcal{L}^{obj}$, *and* $BB \in \mathcal{L}^{flat}$. *Then* $CEW(I)(\varphi | KB \wedge BB) = CEF(I)(\varphi | KB \wedge BB)$.

Thus, CEF and CEW agree in contexts where both are defined.

By analogy to the definition for CEW, we define

$$\Pr_\infty^{CEF}(\varphi | KB \wedge BB) = \lim_{\tau \to 0} \lim_{N \to \infty} CEF(\Pr_N^{\tau, rw})(\varphi | KB \wedge BB).$$

It immediately follows from Theorem 3.4 that

**Corollary 3.5:** If $KB, \varphi \in \mathcal{L}^{obj}$, and $BB \in \mathcal{L}^{flat}$, then $\Pr_\infty^{CEW}(\varphi | KB \wedge BB) = \Pr_\infty^{CEF}(\varphi | KB \wedge BB)$.

As the notation suggests, we view $\Pr_\infty^{CEF}$ as the extension of $\Pr_\infty^{rw}$ obtained by applying CEF. Why did we not define $\Pr_\infty^{CEF}$ as $CEF(\Pr_\infty^{rw})$? Clearly $CEF(\Pr_\infty^{rw})$ and $\Pr_\infty^{CEF}$ are closely related. Indeed, if both are defined, then they are equal.

**Theorem 3.6:** *If both* $CEF(\Pr_\infty^{rw})(\varphi | KB \wedge BB)$ *and* $\Pr_\infty^{CEF}(\varphi | KB \wedge BB)$ *are defined then they are equal.*

It is quite possible, in general, that either one of $\Pr_\infty^{CEF}$ and $CEF(\Pr_\infty^{rw})$ is defined while the other is not. The following example demonstrates one type of situation where $\Pr_\infty^{CEF}$ is defined and $CEF(\Pr_\infty^{rw})$ is not. The converse situation typically arises only in pathological examples. In fact, as we show in Theorem 3.8, there is an important class of cases where the existence of $CEF(\Pr_\infty^{rw})$ guarantees that of $\Pr_\infty^{CEF}$.

**Example 3.7:** Suppose KB is $\|Fly(x) | Bird(x)\|_x \approx 1 \wedge Bird(Tweety)$ and BB is $\Pr(Fly(Tweety) = 0) \wedge \Pr(Red(Tweety) = 1)$. Then, just as we would expect, $\Pr_\infty^{CEF}(Red(Tweety) | KB \wedge BB) = 1$. On the other hand, $CEF(\Pr_\infty^{rw})(Red(Tweety) | KB \wedge BB)$ is undefined. To see why, let $\mu$ be the probability distribution on the four atoms defined by $Fly(Tweety)$ and $Red(Tweety)$ determined by $\Pr_\infty^{rw}(\cdot | KB)$. Since $\Pr_\infty^{rw}(Fly(Tweety) | KB) = 1$, it must be the case that $\mu(Fly(Tweety)) = 1$ (or, more accurately, $\mu(Fly(Tweety) \wedge Red(Tweety)) + \mu(Fly(Tweety) \wedge \neg Red(Tweety)) = 1$). On the other hand, any distribution $\mu'$ over the four atoms defined by $Fly(Tweety)$ and $Red(Tweety)$ that satisfies BB must be such that $\mu'(Fly(Tweety)) = 0$. It easily follows that if $\mu'$ satisfies BB, then $C(\mu', \mu) = \infty$. Thus, there is not a unique distribution over the atoms that satisfies BB and minimizes cross-entropy relative to $\mu$. This means that $CEF(\Pr_\infty^{rw})(Red(Tweety) | KB \wedge BB)$ is undefined. ∎

We next consider what happens when we instantiate CEF with a particular inference process considered by Paris and Vencovska that uses maximum entropy [PV89]. Paris and Vencovska restrict attention to rather simple languages, corresponding to the notion of "essentially propositional" formulas defined below. When considering (our variant) of their method we shall make the same restriction.

We say that $\psi(x)$ is an *essentially propositional* formula if it is a quantifier-free first-order formula that mentions only unary predicates (and no constant or function symbols), whose only free variable is $x$. A *simple knowledge base* KB *about* $c$ has the form $\|\varphi_1(x) | \theta_1(x)\|_x \preceq \alpha_1 \wedge \ldots \wedge \|\varphi_k(x) | \theta_k(x)\|_x \preceq \alpha_k \wedge \psi(c)$, where $\varphi_1, \ldots, \varphi_k, \theta_1, \ldots, \theta_k, \psi$ are all essentially propositional.[8] The ME inference process is only defined for a simple knowledge base about $c$ and an essentially propositional query $\varphi(c)$ about $c$. Let KB $=$ KB$' \wedge \psi(c)$ be an essentially propositional knowledge base about $c$ (where KB$'$ is the part of the knowledge base that does not mention $c$). If the unary predicates that appear in KB are $\mathcal{P} = \{P_1, \ldots, P_k\}$, then KB$'$ can be viewed as putting constraints on the $2^k$ atoms over $\mathcal{P}$.[9] The form of KB$'$ ensures that there will be a unique distribution $\mu_{\text{me}}$ over these atoms that maximizes entropy and satisfies the constraints. We then define ME$(\varphi(c)|$KB$' \wedge \psi(c))$ to be $\mu_{\text{me}}(\varphi|\psi)$. Intuitively, we are choosing the distribution of maximum entropy over the atoms that satisfies KB$'$, and treating $c$ as a "random" element of the domain, assuming it satisfies each atom over $\mathcal{P}$ with the probability dictated by $\mu_{\text{me}}$.

To apply CEF to ME, we also need to put restrictions on the belief base. We say that BB $\in \mathcal{L}^{flat}$ is an *essentially propositional belief base about $c$* if every basic proportion expression has the form $\Pr(\varphi(c)|\theta(c))$, where $\varphi$ and $\theta$ are essentially propositional. (In particular, this disallows statistical formulas in the scope of Pr.) A *simple belief base about $c$* is a conjunction of the form $\Pr(\varphi_1(c)|\theta_1(c)) \le \alpha_1 \wedge \cdots \Pr(\varphi_k(c)|\theta_k(c)) \le \alpha_k$, where all of the formulas that appear are essentially propositional. We can only apply CEF to ME if the knowledge base has the form KB $\wedge$ BB, where KB is a simple knowledge base about $c$ and BB is a simple belief base about $c$. It follows from results of [GHK92, PV89] that random worlds and ME give the same results on their common domain. Hence, they are also equal after we apply the CEF transformation. Moreover, on this domain, if CEF$(\Pr^{rw}_\infty)$ is defined, then so is $\Pr^{\text{CEF}}_\infty$. (The converse does not hold, as shown by Example 3.7.) Thus, we get

**Theorem 3.8:** *If* KB *is a simple knowledge base about $c$,* BB *is a simple belief base about $c$, and $\varphi$ is an essentially propositional formula, then*

$$CEF(ME)(\varphi(c)|\text{KB} \wedge \text{BB}) = CEF(\Pr^{rw}_\infty)(\varphi(c)|\text{KB} \wedge \text{BB}).$$

*Moreover, if CEF(ME)$(\varphi(c)|$KB $\wedge$ BB$)$ is defined, then*

$$CEF(ME)(\varphi(c)|\text{KB} \wedge \text{BB}) = \Pr^{\text{CEF}}_\infty(\varphi(c)|\text{KB} \wedge \text{BB}).$$

### 3.3  RS

The last method we consider, RS, is based on the intuition that degree of belief assertions must ultimately arise from statistical statements. This general idea goes back to work in the field of statistical mechanics [Lan80], where it has been applied to the problem of reasoning about the total energy of physical systems. If the system consists of many particles then what is, in essence, a random-worlds analysis can be appropriate. If the energy of the system is known exactly no conceptual problem arises: some possible configurations have the specified energy, while others are impossible because they do not. However, it turns out that it is frequently more appropriate to assume that all we know is the *expected* energy. Unfortunately, it questionable whether this is really an "objective" assertion about the system in question,[10] and in fact the physicists encounter a problem analogous to that which motivated our paper. Like us, one response they have considered is to modify the assumption of uniform probability and move to maximum entropy (thus using, essentially, an instance of our CEW applied to a uniform prior). But another response is the following. Physically, expected energy is appropriate for systems in thermal equilibrium (i.e., at a constant temperature). But in practice this means that the system is in thermal contact with a (generally much larger) system, sometimes called a *heat bath*. So another approach is to model the system of interest as being part of a much larger system, including the heat bath, whose total energy is truly fixed. On this larger scale, random-worlds is once again applicable. By choosing the energy for the total system appropriately, the expected energy of the small subsystem will be as specified. Hence, we have converted subjective statements into objective ones, so that we are able to use our standard techniques. In this domain, there is a clear physical intuition for the connection between the objective information (the energy of the heat bath) and the subjective information (the expected energy of the small system).

A more recent, and quite different, appearance of this intuition is in the work of Paris and Vencovska [PV92]. They defined their method so that it has the same restricted scope as the ME method. We present a more general version here, that can handle a somewhat richer set of knowledge bases, although its scope is still more restricted than CEF. It can deal with arbitrary inference processes, but the knowledge base must have the form KB $\wedge$ BB, where KB is objective and BB is an essentially propositional belief base about some constant $c$. The first step in the method is to transform BB into an objective formula. Let $S$ be a new unary predicate, representing the set of individuals "just like $c$". We transform BB to KB$_{\text{BB}}$ by replacing all terms of the form $\Pr(\psi(c)|\theta(c))$ by $\|\psi(x)|\theta(x) \wedge S(x)\|_x$, and replacing all occurrences of $\le$ by $\preceq$. We then add the conjuncts $\|S(x)\|_x \approx 0$ and $S(c)$, since $S$ is assumed to be a small set and $c$ must be in $S$. For example, if BB is $\Pr(Red(c)) \le 0.8 \wedge \Pr(Small(c)) = 0.6$, then the corresponding KB$_{\text{BB}}$ is $(\|Red(x)|S(x)\|_x \preceq 0.8) \wedge (\|Small(x)|S(x)\|_x \approx 0.6) \wedge (\|S(x)\|_x \approx 0) \wedge S(c)$. We then define $RS(I)(\varphi(c)|\text{KB} \wedge \text{BB}) = I(\varphi(c)|\text{KB} \wedge \text{KB}_{\text{BB}})$. It is almost immediate from the definitions that if BB is a simple belief base about $c$, then $RS(\Pr^{rw}_\infty)(\varphi(c)|\text{KB} \wedge \text{BB}) = \lim_{\tau \to 0} \lim_{N \to \infty} RS(\Pr^{\tau,rw}_N)(\varphi|\text{KB})$. We abbreviate $RS(\Pr^{rw}_\infty)$ as $\Pr^{\text{RS}}_\infty$.

In general, RS and CEF are distinct. This observation follows from results of [PV92] concerning an inference process CM, showing that RS(CM) cannot be equal to CEF(CM). On the other hand, they show that, in the restricted setting in which ME applies, RS(ME) $=$ CEF(ME). Since ME $= \Pr^{rw}_\infty$ in this setting, we have:

---

[9] An atom over $\mathcal{P}$ is an atom (as defined above) over the formulas $P_1(x), \ldots, P_k(x)$.

[10] If it *is* objective, it is most plausibly a statement about the average energy over time. While this is a reasonable viewpoint, it does not really escape from philosophical or technical problems either.

**Theorem 3.9:** *If* KB *is a simple knowledge base about* $c$, BB *is an essentially propositional knowledge base about* $c$, *and* $\psi$ *is an essentially propositional formula, then*

$$CEF(\Pr^{rw}_{\infty})(\varphi(c)|\text{KB} \wedge \text{BB}) = CEF(ME)(\varphi(c)|\text{KB} \wedge \text{BB})$$
$$= RS(ME)(\varphi(c)|\text{KB} \wedge \text{BB}) = \Pr^{RS}_{\infty}(\varphi(c)|\text{KB} \wedge \text{BB}).$$

## 4  Discussion

We have presented three methods for extending inference processes so that they can deal with degrees of belief. We view the fact that the three methods essentially agree when applied to the random-worlds method as evidence validating their result as the "appropriate" extension of random worlds.

Since our focus was on extending the random-worlds method here, there were many issues that we were not able to investigate thoroughly. We mention two of the more significant ones here:

- Our definitions of CEF and RS assume certain restrictions on the form of the knowledge base, which are not assumed in CEW. Is it possible to extend these methods so that they apply to more general knowledge bases? In this context, it is worth noting that RS has quite a different flavor from the other two approaches. The basic idea involved seems to be to ask "What *objective* facts might there be to cause one to have the beliefs in BB?". Given an answer to this, we add these facts to KB in lieu of BB; we can then apply whatever inference process we choose. We do not see any philosophical reason that prevents application of this idea in wider contexts than belief bases about some constant $c$. The technical problems we have found trying to do this seem difficult but not deep or intractable.

- We have essentially assumed what might be viewed as concurrent rather than sequential updating here. Suppose our knowledge base contains two constraints: $\Pr(\varphi_1) = \alpha_1 \wedge \Pr(\varphi_2) = \alpha_2$. Although we cannot usually apply Jeffrey's rule to such a conjunction, we can apply the rule sequentially, first updating by $\Pr(\varphi_1) = \alpha_1$, and then by $\Pr(\varphi_2) = \alpha_2$. We have described our methods in the context of updating by any set of constraints at once, but they can also be defined to update by constraints one at a time. The two possibilities usually give different results. Sequential updating may not preserve any but the last constraint used, and in general is order dependent. Whether this should be seen as a problem depends on the context. We note that in the very special case in which we are updating by objective facts (i.e., conditioning) sequential and concurrent updating coincide. This is why this issue can be ignored when doing Bayesian conditioning in general, and in ordinary random-worlds in particular. We have only considered concurrent updates in this paper, but the issue surely deserves deeper investigation.